%% file: main.tex
\documentclass[sigconf,10pt]{acmart}  
\usepackage{times}
\usepackage{soul}
\usepackage{url}
\usepackage[utf8]{inputenc}
\usepackage{caption}
\usepackage{tikz}
\usepackage{amsmath,amsfonts, bm}
\usepackage{algorithm}
\usepackage{algorithmicx}
\usepackage{algpseudocode}

\usepackage{graphicx}
\usepackage{textcomp}
\usepackage{booktabs} 
\usepackage{siunitx}
\usepackage{xspace,url}

\usepackage{color}
\usepackage{colortbl}
\usepackage{multirow}
\usepackage{multicol}
\usepackage{makecell}
\usepackage{tabularx}
\usepackage{listings}
\usepackage{graphicx}
\usepackage{caption}
\usepackage{enumitem}
\usepackage{adjustbox}
\usepackage{subcaption}

\usepackage[colorinlistoftodos]{todonotes}

\acmYear{2024}\copyrightyear{2024}
\setcopyright{rightsretained}
\acmConference[ACM MobiCom '24]{International Conference On Mobile Computing And Networking}{September 30--October 4, 2024}{Washington D.C., DC, USA}
\acmBooktitle{International Conference On Mobile Computing And Networking (ACM MobiCom '24), September 30--October 4, 2024, Washington D.C., DC, USA}
\acmDOI{10.1145/3636534.3649379}
\acmISBN{979-8-4007-0489-5/24/09}

\definecolor{airforceblue}{rgb}{0.36, 0.54, 0.66}
\newcommand{\shpd}[1]{\textcolor{black}{#1}}
\newcommand{\wh}[1]{\textcolor{black}{#1}}    

\newcommand{\name}{AutoDroid\xspace} 
\newcommand{\datasetname}{DroidTask\xspace}

\newcommand{\ie}{\textit{i}.\textit{e}.~}
\newcommand{\eg}{\textit{e}.\textit{g}.~}

\author{Hao Wen$^{1}$, Yuanchun Li$^{1,2,\dagger}$, Guohong Liu$^{1}$, Shanhui Zhao$^{1,*}$, Tao Yu$^{1,*}$, \\Toby Jia-Jun Li$^{3}$, Shiqi Jiang$^{4}$, Yunhao Liu$^{5}$, Yaqin Zhang$^{1}$, Yunxin Liu$^{1,2}$}
\thanks{$\dagger$ Corresponding author: Yuanchun Li (liyuanchun@air.tsinghua.edu.cn).}
\thanks{* Shanhui Zhao and Tao Yu were student interns at Tsinghua University.}
\affiliation{$^1$ Institute for AI Industry Research (AIR), Tsinghua University\\ 
$^2$ Shanghai Artificial Intelligence Laboratory\\ 
$^3$ Department of Computer Science and Engineering, University of Notre Dame\\ 
$^4$ Microsoft Research\\ 
$^5$ Global Innovation Exchange \& Department of Automation, Tsinghua University}

\renewcommand{\shortauthors}{Wen and Li et al.}

\begin{document}

\title{\name: LLM-powered Task Automation in Android}




\begin{abstract}
Mobile task automation is an attractive technique that aims to enable voice-based hands-free user interaction with smartphones.
However, existing approaches suffer from poor scalability due to the limited language understanding ability and the non-trivial manual efforts required from developers or end-users.
The recent advance of large language models (LLMs) in language understanding and reasoning inspires us to rethink the problem from a model-centric perspective, where task preparation, comprehension, and execution are handled by a unified language model. 
In this work, we introduce \name, a mobile task automation system \shpd{capable of handling} arbitrary tasks on any Android application without manual efforts. 
The key insight is to combine the commonsense knowledge of LLMs and domain-specific knowledge of apps through automated dynamic analysis.
The main components include a functionality-aware UI representation method that bridges the UI with the LLM, exploration-based memory injection techniques that augment the app-specific domain knowledge of LLM, and a multi-granularity query optimization module that reduces the cost of model inference. 
We integrate \name with off-the-shelf LLMs including online GPT-4/GPT-3.5 and on-device Vicuna, and evaluate its performance on a new benchmark for memory-augmented Android task automation with 158 common tasks. The results demonstrated that \name is able to precisely generate actions with an accuracy of 90.9\%, and complete tasks with a success rate of 71.3\%, outperforming the GPT-4-powered baselines by 36.4\% and 39.7\%. 
\end{abstract}

\ccsdesc[500]{Human-centered computing~Ubiquitous and mobile computing}
\ccsdesc[300]{Computing methodologies~Artificial intelligence}

\keywords{Task Automation, Large Language Models, App Analysis}

\maketitle

\input{tex/intro}

\input{tex/background}
\input{tex/approach}
\input{tex/implementation}
\input{tex/experiment}

\input{tex/related_work}
\vspace{-0.25cm}
\input{tex/discussion}
\vspace{-0.25cm}
\input{tex/conclusion}

\balance
\bibliographystyle{ACM-Reference-Format}
\bibliography{reference}

\end{document}

%% file: tex/intro.tex
\section{Introduction}



Smartphone is one of the most sophisticated devices for individuals. With millions of mobile applications (apps for short) that have access to various embedded sensors and rich personal data, smartphones \shpd{can be used} for a lot of daily tasks such as ordering food, managing social networks, sensing and tracking health conditions, etc. 
Therefore, how to intelligently automate tasks on smartphones has become an attractive topic for mobile developers and researchers, due to its potential to significantly improve user experience and enable helpful virtual personal assistants.


The major approaches to mobile task automation can be classified as developer-based, demonstration-based, and learning-based techniques.
Most existing commercial products (\eg Siri, Google Assistant, Cortana, etc.) take a developer-based approach, which requires significant development efforts to \shpd{support} a new task.
For example, to enable an automated task with Google Assistant,
app developers need to identify the functionality which they want to trigger, configure and implement the corresponding intent, and register the intent \shpd{with} the assistant. When executing a task, the assistant uses natural language understanding (NLU) modules to map the user command to the intent, extract the intent parameters, and invoke the corresponding developer-defined function.
Researchers have explored various methods to ease the development efforts. However, these methods still suffer from poor scalability, since they either require ad-hoc and/or large-scale human demonstrations of tasks (\eg programming-by-demonstration approaches \cite{kite, ulink,li2017programming_iot} and supervised learning approaches \cite{seq2act, metagui, motif}) or require defining a clear reward for task completion (\eg reinforcement learning approaches \cite{ToyamaEtAl2021AndroidEnv, rl_use_computer, glider}).
Due to the lack of scalability, there are few automated tasks supported today, even in the most popular apps.


\begin{figure}
    \centering
    \includegraphics[width=0.47\textwidth]{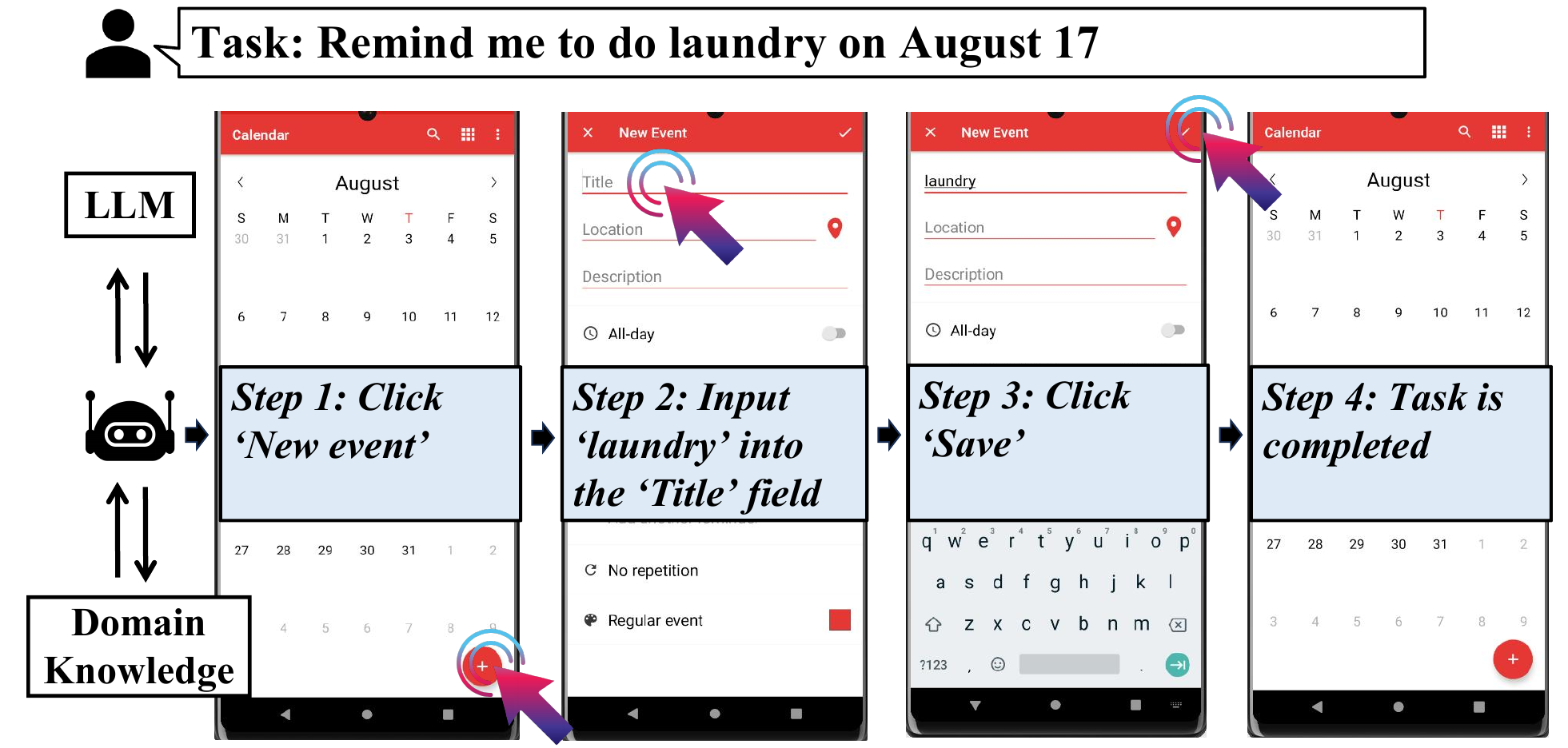}
    \caption{An illustration of LLM-powered mobile task automation. 
    The agent interacts with the smartphone GUI to complete an arbitrary task, with the guidance of LLM and app domain knowledge.}
    \label{fig:intro}
    \vspace{-0.5cm}
\end{figure}

Recently, the emergence of large language models (LLMs) like ChatGPT \cite{chatgpt} and Claude \cite{claude} shows the promise \shpd{in solving} the scalability issue of task automation.
Compared to traditional models, LLMs demonstrate unique abilities such as instruction following \cite{alpaca}, step-by-step reasoning \cite{cot}, and zero-shot generalization \cite{zero-shot-cot}.
Such abilities are enabled by self-supervised learning on a huge corpus (more than 1.4 trillion tokens \cite{touvron2023llama}) followed by tuning with human feedback.
With these capabilities, researchers have managed to let LLMs invoke tools automatically, such as search engines \cite{webgpt}, code interpreters \cite{copilot}, and third-party APIs \cite{gorilla, chamelon}. 
Similarly, using LLMs can potentially avoid the cumbersome manual efforts in mobile task automation. Meanwhile, connecting LLMs to smartphones can further unleash the power of LLMs in personal domains.



The goal of LLM-powered mobile task automation is to build an autonomous agent that can complete user-specified tasks by interacting with the smartphone. \wh{Although existing research \cite{talking_with_ui} attempts to \shpd{enable} LLMs to understand mobile UIs, it simply relies on prompt engineering and \shpd{does} not utilize app-specific domain knowledge. \name combines the capabilities of LLM and the app-specific knowledge through dynamic app analysis, which enables it to handle arbitrary unseen tasks without manual efforts (illustrated in Figure~\ref{fig:intro}). } 
We identify three key problems to achieve this goal.
\begin{enumerate}
    \item \textbf{GUI Representation.} The input and output of task automators are graphical user interface (GUI) states and actions, unlike the natural language sentences LLMs can handle.
    To help LLMs better understand the GUI information and make precise interaction decisions, the GUI states and actions must be converted to text format while incorporating rich structured information.
    \item \textbf{Knowledge Integration.} Solving tasks with LLMs requires domain-specific knowledge about the applications.
    Unlike other tools studied in prior work (\eg APIs) that LLMs can be easily configured to use, a smartphone app is usually a more complicated automata. LLMs need to navigate between different states to figure out how to complete the tasks.
    \item \textbf{Cost Optimization.} Querying LLMs is costly and compute-intensive, while completing a task with LLMs may involve many lengthy queries due to complexity of tasks and smartphone apps. Thus, it is desirable to optimize the efficiency of LLM queries to facilitate responsive task automation experience.
\end{enumerate}


We introduce a mobile task automation framework, \name, to \shpd{address} the above problems.
Overall, \name executes tasks by prompting the LLMs with an HTML-style text representation of GUI and querying for action guidance. To augment the LLMs with app knowledge, \name randomly explores the target apps and \shpd{extracts} UI transition graphs from them. By analyzing the UI states and transitions with LLMs, \name can \shpd{convert} the raw information to task completion knowledge, which is then integrated into the task automator by injecting foreseen functionalities into the prompts, matching relevant UI traces, or tuning the LLM parameters.
The cost of querying LLMs is reduced by reducing and simplifying the queries based on app knowledge.

To systematically study the performance and challenges of LLM-powered task automation on Android, we build a benchmark with 158 manually labeled tasks from 13 open-source common mobile apps (Calendar, Messenger, Contacts, etc.).
The source code and executable environments of the apps are provided for obtaining auxiliary information and reproducing task executions.
The tasks include frequently asked how-to questions from the PixelHelp \cite{seq2act} dataset and common functionalities in the apps.
For each task, we manually labeled the steps to complete the tasks, where each step is associated with both the GUI state and the GUI action.
Our benchmark evaluates the performance of LLM-powered task automation in terms of accuracy and cost.

We evaluate the effectiveness of our \name approach on the benchmark with different types of LLMs, including state-of-the-art online LLM services (GPT-3.5 and GPT-4) and open-source on-device LLMs (Vicuna). 
The results have demonstrated that \name can complete unseen tasks with a success rate of 71.3\% with GPT-4, in which each action is selected with an accuracy of 90.9\%.
As compared with the baselines powered by off-the-shelf LLMs, the task completion rates are improved by 36.4\% to 39.7\%, and the average cost of querying LLMs is reduced by 51.7\%. 

Our work makes the following technical contributions:
\begin{enumerate}
    \item To the best of our knowledge, this is the first work on enhancing mobile task automation by combining LLMs and app-specific knowledge. We build a benchmark for this problem.
    \item We introduce a novel UI representation method that connects smartphones with LLMs, a task synthesis method \shpd{for augmenting} LLMs with app knowledge, and various LLM query optimization techniques to reduce the cost of task automation.
    \item Through a comprehensive evaluation, we demonstrate the effectiveness of our \shpd{approach} and the potential to advance the field of mobile task automation.
\end{enumerate}

%% file: tex/background.tex
\section{Background and Motivation}
\label{section:background}


\subsection{Mobile Task Automation}


The goal of mobile task automation is to automatically complete different kinds of tasks given by users. Its input is an arbitrary task described with natural language and a mobile app to execute the task. The output is a sequence of UI actions that can be executed on a smartphone. 



A \textbf{task} is a multi-step functionality request from the user intended for completion on a smartphone, often lacking explicit instructions. 
A \textbf{UI state}, visible to users on their mobile device, is an arrangement of controls depicted through images and text, typically organized as a GUI tree. 
A \textbf{UI action}, performable by the user or an agent on the device's screen, is defined by a tuple \textit{(target element, action type, value)}. \textit{Target element} refers to a control in the UI state, such as a button, text box, input field, or slider. \textit{Action type} represents how the target element is manipulated. We consider three main types of smartphone interactions, including ``click'', ``input'', and ``swipe''. The \textit{value} field is the text content of the ``input'' action, which is empty for other action types.



\wh{
In contrast to existing methods that utilize LLMs to summarize or \shpd{respond} to queries about individual mobile UIs \cite{talking_with_ui, venkatesh2022ugif}, automating mobile tasks demands the capability to plan task solutions and an in-depth understanding of which UIs are essential for task completion. \name aims to achieve multi-step task automation by leveraging app-specific knowledge. 
Furthermore, unlike most existing approaches that require significant developer/user efforts \cite{ulink} to enable automated tasks, we aim to achieve unsupervised task automation, \ie support the automation of arbitrary tasks on black-box apps (whose internal mechanisms are unknown) without human effort. }
However, we assume that the apps are available for automated analyses, \eg exploring the states, crawling the content, and analyzing the code. Such an assumption is reasonable because the app packages are all available for download and static/dynamic app analysis techniques have been extensively studied before \cite{droidbot, Caiipa, li2019humanoid, VDfarms}.
\vspace{-0.13cm}
\subsection{Large Language Models}
Large language models (LLMs for short) mainly refer to the Transformer-based \cite{transformer} language models that contain billions of parameters and are trained on massive amounts of text data, such as ChatGPT \cite{chatgpt}, GPT-4 \cite{openai2023gpt4}, PaLM \cite{palm}, LLaMA \cite{touvron2023llama}, etc. 
These models exhibit capabilities that are not present in smaller models, 
including mathematical reasoning \cite{solve_math}, program synthesis \cite{copilot}, and multi-step reasoning \cite{cot}. Specifically, LLM can perform the tasks better than the benchmark models trained on dedicated datasets. 
The input of an LLM is a prompt, which is an instruction to guide its generation of responses. The prompt is tokenized into tokens (words or subwords) before being fed into the LLM.

Researchers are actively exploring methods to enhance the problem-solving capabilities of LLMs by incorporating reasoning skills \cite{cot} and tool utilization \cite{chamelon, gorilla, responsible_task_automation}. These efforts aim to enable LLMs to use tools by teaching them to call APIs or to synthesize codes. However, task automation in smartphone apps is more complex since it is often related to the environment without documented interfaces. 


\subsection{LLM meets Mobile Task Automation}


We believe that incorporating LLMs into mobile task automation brings unique advantages and strengths to both fields.


First, \textbf{LLMs have the potential to significantly advance the applications of mobile task automation.}
The voice-controlled intelligent personal assistants (IPA) are typical applications of mobile task automation, \shpd{aiming} to provide intelligent, efficient, hands-free user experience on mobile devices.
Such applications are not only useful in smartphones, but also in many other scenarios, including automotive in-vehicle infotainment (IVI) systems \cite{IVIsys}, wearable fitness trackers \cite{uiwear, iself}, and VR/AR devices \cite{AMash}.
To support IPA services, developers usually have to manually configure the task workflows, which is a cumbersome process even for experienced developers.
Researchers have also attempted to build agents that can directly manipulate GUI elements like human users \cite{seq2act, metagui, kite, talking_with_ui}. 
However, they usually require a lot of human demonstrations, step-by-step instructions, or clearly-designed task-specific reward functions for task completion \cite{rl_use_computer, glider}. 
LLM-based agents can be better at GUI task automation with their strong language comprehension and reasoning abilities.


Second, \textbf{equipping LLMs with smartphones can significantly augment their abilities.}
LLMs are trained with large-scale public data that contains rich commonsense and world knowledge, while they have limited knowledge about individual users and limited abilities to provide personalized services.
Smartphones have been an important part of daily life by helping people connect with others, stay organized with calendars, navigate and get directions, control smart-home devices, and so on.
If LLMs learn to use smartphone apps and access data siloed in them, they could become much better personal assistants with access to the rich sensors and personal data in mobile apps.
\begin{figure}
    \centering
    \includegraphics[width=0.37\textwidth]{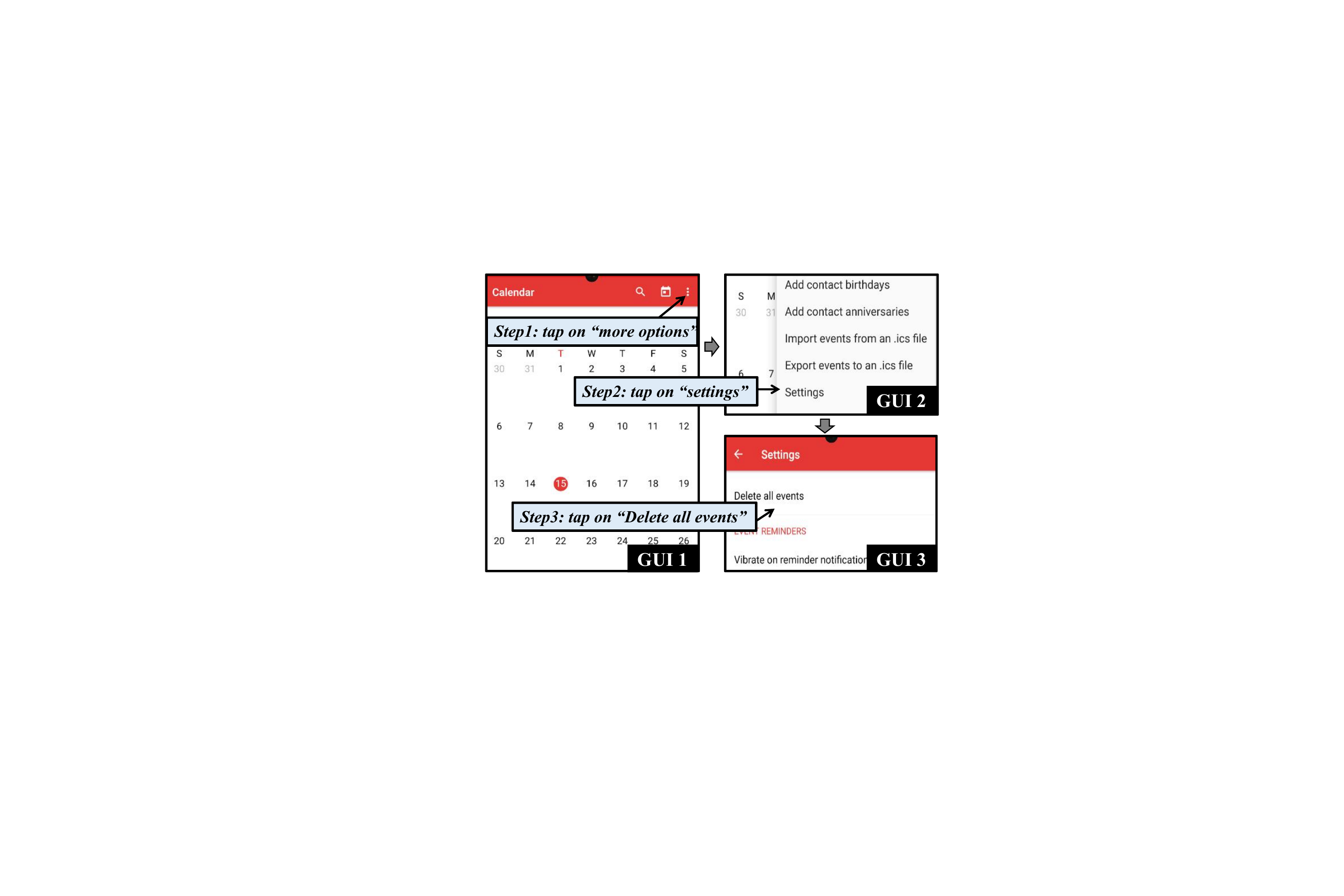}
    \caption{An example task of \textit{``Remove all the events in the Calendar''}. The agent needs to tap on \textit{"more options"} and \textit{"settings"} on the first and second GUI, which do not have a direct semantic association with the ultimate goal. This association can be grasped more \shpd{easily} with app analysis.}
    \label{fig:motivation}
    \vspace{-0.2cm}
\end{figure}

Yet, applying LLMs to mobile task automation involves several challenges, including GUI representation, knowledge integration, and cost optimization. 
First, LLMs are only capable of processing plain text data and cannot directly handle GUI or interact with it. Although the GUI state in Android can be represented as text using the UI Hierarchy Viewer or Accessibility Services, it is usually lengthy (about 40k tokens on average for each UI state) and difficult for LLMs to interpret.
Second, LLMs lack knowledge and experience about certain applications, which may lead to incorrect execution of instructions. Figure \ref{fig:motivation} shows an example where a deep understanding of the app is needed to complete the task. It is difficult to determine solely based on semantics and prior knowledge that clicking on \textit{`more options'} and then \textit{'settings'} on the first two screens will lead to the screen containing the option to \textit{`delete all events'}.
\wh{Therefore, relying \shpd{solely} on prompt engineering for LLMs to produce common-sense solutions can result in mistakes. A better approach might be to let LLMs investigate and learn from mobile apps, gaining practical experience prior to undertaking tasks for users. }
Third, using LLMs for task completion may be costly. The price of querying ChatGPT API \cite{chatgpt} is \$1.5 / 1000K tokens. Even if we can deploy a private LLM service, the computational cost is still high. For example, inferring a single token with LLaMA-7B \cite{touvron2023llama} takes 6.7 billion FLOPs,
and the whole process of task completion may use over 2000 tokens.

%% file: tex/approach.tex
\section{Our Approach: \name}
\label{sec:approach}

\begin{figure*}
    \centering
    \includegraphics[width=0.8\textwidth]{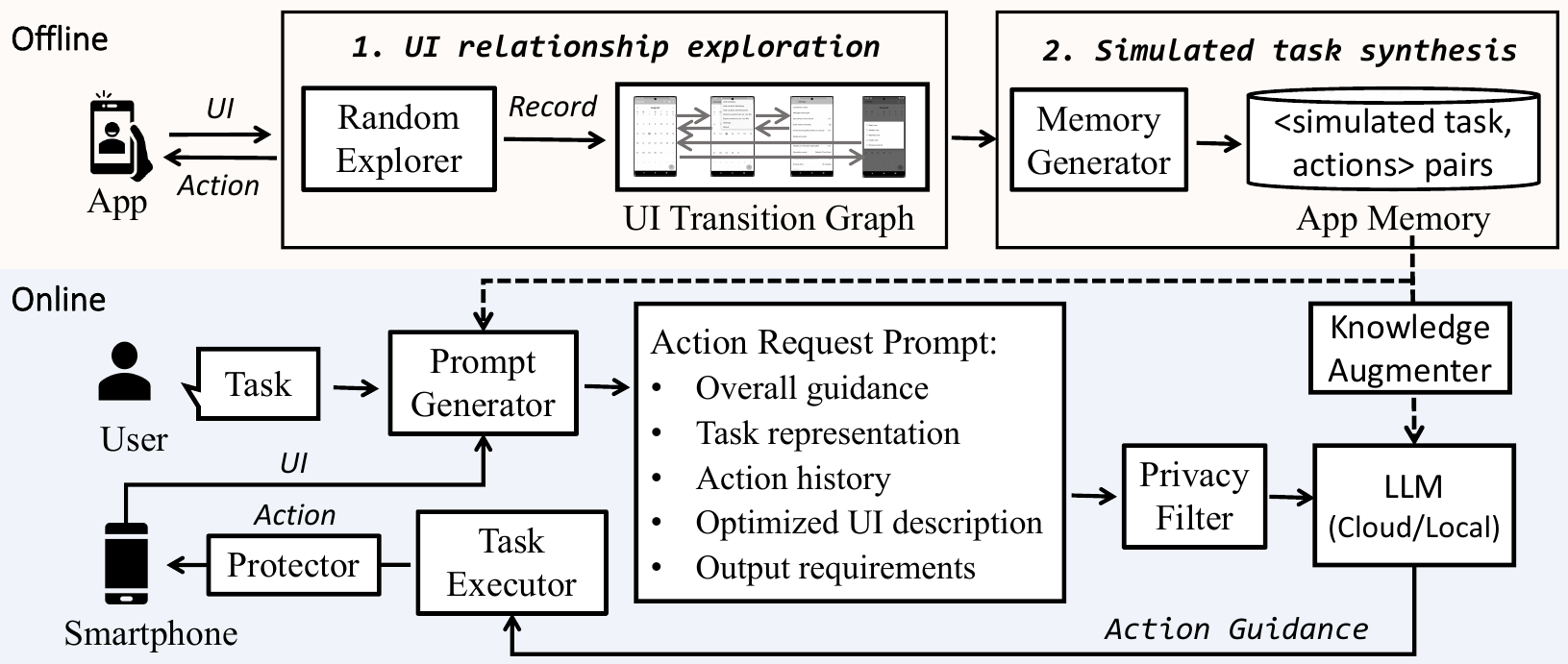}
    \vspace{-0.3cm}
    \caption{The workflow of \name.}
    \label{fig:overview}
    \vspace{-0.2cm}
\end{figure*}

We introduce \name, an LLM-powered end-to-end mobile task automation system to solve the aforementioned challenges. In the offline stage, \name obtains app-specific knowledge by exploring UI relations and synthesizing simulated tasks. In the online stage, \name continuously queries the memory-augmented LLMs to obtain guidance on the next action. The task is completed by following the LLM-suggested actions. \name adopts several techniques to improve the task completion rate and optimize the query cost. Figure \ref{fig:overview} illustrates the workflow. 


We explain the functioning of \name using the example of automating tasks in a calendar app:
During the offline stage, \name explores the app by randomly clicking buttons on the screen and records the result in a UI Transition Graph (UTG) memory (\textit{Step 1}). Next, it traverses all the UI elements in the UTG and summarizes the tasks they can accomplish (\textit{Step 2}).
During online operation, when the user issues a command such as ``delete all the events in the calendar'', the \textit{Prompt Generator} generates a prompt based on the task, the UI state description, and relevant information stored in the \textit{App Memory}. This information includes instructions on how to navigate to the GUI page that contains the ``delete events'' option.
Subsequently, the \textit{Privacy Filter} replaces any sensitive information in the prompt to safeguard privacy. The filtered prompt is then sent to the LLM.
Once the LLM provides an answer, the \textit{Task Executor} parses the action that can be executed on the smartphone and verifies its security before performing it. If the executor deems the action to be potentially risky, such as ``delete all the events'' in this particular task, it will seek confirmation from the user before proceeding. We will explain how \name does all of these in the rest of this section.

\subsection{Task-oriented UI Prompting}
\label{section: prompting}

\begin{figure}
    \centering
    \includegraphics[width=0.45\textwidth]{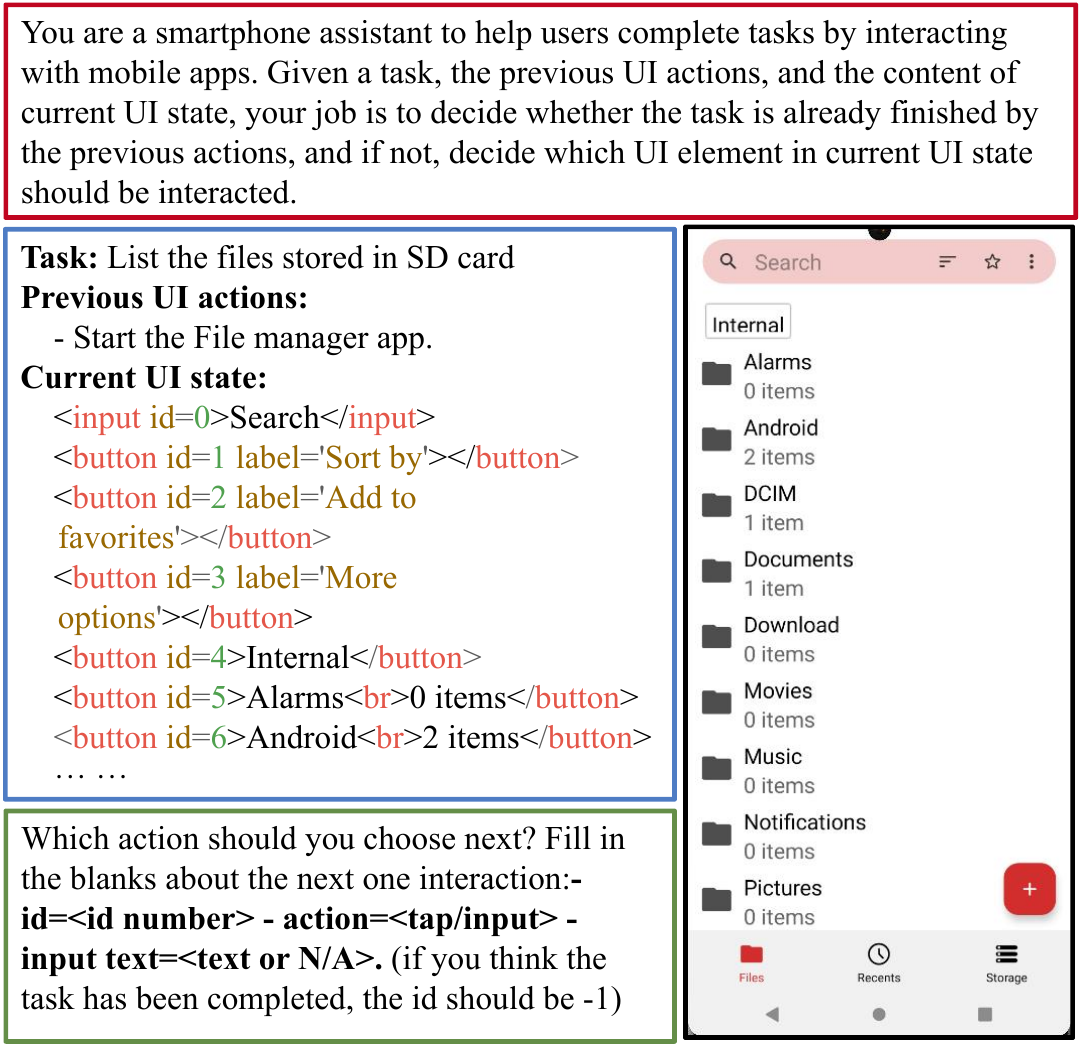}
    \vspace{-0.3cm}
    \caption{An illustration of the prompt used by \name. The content in red, blue, and green boxes are the overall guidance, the task representation, and the output requirements respectively. The `current UI State' in the prompt refers to the UI displayed within the black box.}
    \label{fig:prompt}
    \vspace{-0.2cm}
\end{figure}

UI prompting refers to the process of representing underlying UI information in text and injecting it into the prompt to query the LLM. 
The goal of UI prompting is to clearly present the UI textual and structural content to the LLM and restrict the output of the LLM to predict only valid UI interactions. 
Figure \ref{fig:prompt} showcases an example of \name converting a GUI interface into a prompt \shpd{while} completing the task.

\subsubsection{Converting GUI to Simplified HTML Representation} 
\label{sec:html_desc}
We develop a GUI parsing module to convert GUI to a simplified HTML representation that can be processed by LLMs. Researchers have found that LLMs are better at understanding HTML than natural-language-described UIs due to the large amount of HTML code in the training data of LLMs \cite{talking_with_ui}. 
Therefore, we represent the GUI in HTML style, which can preserve the attribute information of UI elements. We use five types of HTML tags, namely <button>, <checkbox>, <scroller>, <input>, and <p>, which represent elements that can be clicked, checked, swiped, edited, and any other views respectively. The properties included for each element are: ID (the order in which the element appears in the GUI tree), label (the content description that describes the function of the element), onclick (hints about the UI states that will be accessed upon clicking this button or checking/unchecking this checkbox, which will be introduced in \S\ref{sec:prompt_augment}), text (the text on the element), direction (scrolling direction, including up/down/left/right), checked (whether the checkbox is checked or not), value (the text that has been input to the text box). The classes and properties of GUI elements are shown in Table \ref{tab:element_properties}. 
We further simplify the DOM tree by pruning invisible elements and merging functionally equivalent elements, which will be introduced in \S \ref{section: query_optimization}. The texts of two merged UI elements are separated by ``<br>'', which represents a line break in HTML.

\begin{table}
	\caption{The classes and properties of GUI elements}
	\vspace{-0.3cm}
	\centering
        \resizebox{.47\textwidth}{!}
        {
	\begin{tabular}{lll}
		\toprule
		 Class                     & Properties                                        & Available action\\
		\midrule
		<button>                & ID, label, onclick (\S\ref{sec:prompt_augment}), text                      & click \\
		<checkbox>              & ID, checked, label, onclick (\S\ref{sec:prompt_augment}), text              & check/uncheck\\
		<scroller>              & ID, scroll direction, label, text     & scroll <direction>\\
		<input>                 & ID, label, text, value                & input <text>\\
		<p>                     & ID, label, text                       & N/A\\  
		\bottomrule
	\end{tabular}
 }
	\label{tab:element_properties}
     \vspace{-0.5cm}
\end{table}

In our experiments, we observe that the agent generally does not proactively scroll on interfaces that can be scrolled vertically (shown in \S \ref{eval:action_selection}). However, having information about the scrolled interface is crucial for decision-making, especially when the target button is located on a scrolled portion of the interface that is not yet visible. Therefore, to provide the agent with comprehensive information, we need to include the components from the scrolled portion of the interface in the current UI state. 
To achieve this, for a given interface, \name first automatically scrolls through all scrollable components and records the information of the visible UI elements, and then provides this information to the LLM for decision-making. This approach offers two advantages. Firstly, it prevents LLMs from making blind selections when they cannot see all the information on the interface. Secondly, it eliminates the need for LLMs to provide explicit instructions for scrolling, reducing the frequency of calling the LLM and lowering the associated computational overhead.


\subsubsection{Restricting the Action Space with Selections}
A key characteristic of UI task automation is that all agent actions need to be confined to the constraints of the underlying app \textit{i.e.}, the agent can only perform actions of a supported action type on one of the existing UI elements. Thus, a challenge is to adapt LLMs, which are generative in nature, to such a discrete choice task. 
Hence, we impose the necessity for LLMs to produce results in a predetermined structure by completing the following requirement: ``- id=<id number> - action=<tap/input> input text=<text or N/A> (in the event of task completion, id=-1)''. LLMs must refrain from generating id or input in an arbitrary format.

\subsection{Exploration-based Memory Injection}

Exploration-based memory injection aims to provide app-related information to LLMs, enabling them to gain insights into apps, understand app utilization methods, and make effective decisions. However, there are challenges in utilizing automated app-related knowledge to assist LLMs in task automation, including: (i) The UI Transition Graph (UTG) obtained through random exploration cannot be directly processed by the LLM. (ii) Memory acquired solely through UI automation tools contains only UI and action data, without the essential information needed to directly enable task automation. This includes details about the specific UI elements and actions necessary to accomplish a particular task. 
(iii) An app may have numerous UI screens and UI elements (buttons, text boxes, etc.), exceeding the token length limit of LLMs if all of them are included in a prompt.
To overcome these challenges, \name synthesizes simulated tasks based on the randomly explored UI graph. 
These simulated tasks serve as a guide for LLMs on how to accomplish a user task.


\subsubsection{Simulated Task Generation}

\begin{figure*}
    \centering
    \includegraphics[width=0.9\textwidth]{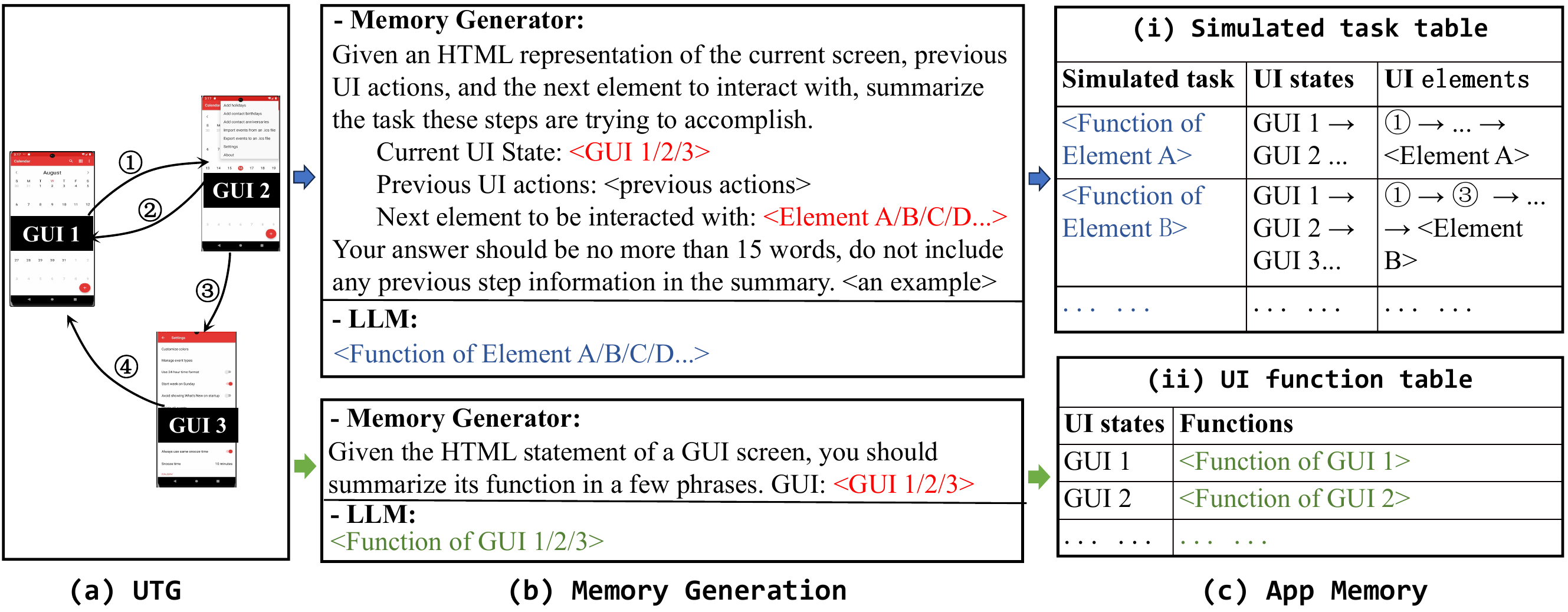}
    \caption{Workflow of offline simulated task synthesis. Given the UI Transition Graph (UTG), Memory Generator synthesizes a simulated task for each UI element with LLMs, \shpd{then} records the task-states-elements in the App Memory.}
    \label{fig:generating_functions}
    \vspace{-0.4cm}
\end{figure*}

\name generates simulated tasks by analyzing the UI Transition Graph (UTG) as depicted in Figure \ref{fig:generating_functions}. 
The UTG generated by the UI automator contains crucial information about the application, such as the connections between UIs and the presence of different UI elements on each screen. By summarizing the functionalities of all UI elements, we can gain a thorough understanding of the tasks that can be performed within the app and determine the corresponding UI elements required to execute them. As a result, \name parses all UI states and UI elements present in the UTG and extracts their functions by querying LLMs.

Specifically, UTG can be regarded as a directed graph, where the nodes and edges are all UI states and actions recorded by the random Explorer, denoted as $\mathbf{U}$ and $\mathbf{A}$ respectively. 
For each UI state $\mathbf{U}_i$, the memory generator queries LLM to summarize the functionalities of all the UI elements $\{e_i^{j}\}_{j=1}^{|U_i|}$, where $|U_i|$ denotes the number of elements in $U_i$. 
Note that \name only extracts the functionality of an element on the UI state that is closest to the initial UI if it appears on multiple UI states.
After traversing all UI elements in the UTG, we obtain the \textit{simulated task table} in the app memory containing $n$ entries, where $n$ represents the total number of UI elements on the UTG. Each entry in the table corresponds to a UI element $e_i^{j}$ and is divided into three parts: \textit{<Simulated task, UI states, UI elements>}.  ``Simulated task'' represents the functionality of $e_i^{j}$ that has been summarized by LLM, which can be perceived as a simulated task that can be completed by clicking this element. ``UI elements'' includes all the elements that were clicked, starting from the initial UI of the app and leading up to the attainment of $U_i$. ``UI states'' represents the sequence of UI states that were traversed from the initial UI state to $U_i$. 
This table provides the agent with information about the required operations to achieve each functionality, aiding the agent in planning how to complete a given task efficiently.
Apart from the \textit{simulated task table}, there is an additional table called the \textit{UI function table} in the app memory. It provides a summary of the functionality associated with each UI state in the UTG. This information is obtained by querying the LLM to summarize the function of each UI state.

\subsubsection{Augmenting Prompts with App Memory}
\label{sec:prompt_augment}


The most straightforward approach to leveraging app-specific knowledge is to incorporate the app memory directly into the prompt, which can provide guidance to the LLM.
However, this may exceed the maximum token limit of the LLM such as 4097 tokens for GPT-3.5 \cite{chatgpt}. 
In many cases, only a few UI elements are necessary to complete a user's task.
Hence, we selectively incorporate the most relevant UI information into the prompt.

\name determines the importance of a UI element in the app memory based on the similarity between its simulated task and the current user task.
We use an embedding model (Instructor-XL \cite{instructor-xl}) that maps natural language sentences to a fixed-dimension embedding, where the embeddings of sentences with similar meanings are closer. The cosine similarity between the embeddings of the simulated task $S$ and the current task $T$ is denoted as $sim(E(S), E(T))$.
Then, we can find k most similar simulated tasks in the app memory, denoted as $\{S_1, S_2,..., S_k\}$.
For each $S_i$, we can retrieve the corresponding \textit{UI states} and the \textit{UI elements} from the ``simulated task table'' in the app memory. 
In the online stage, if the current UI matches one of the \textit{UI states} associated with $S_i$, we give hints about the UI elements that the random explorer interacted with in this UI state. This helps LLMs understand the outcome of interacting with the elements. 
Specifically, the prompt generator of \name will add a new property ``onclick'' to the HTML UI statement in the prompt (shown in Table\ref{tab:element_properties}). In HTML, ``onclick'' is used to describe the event that will occur when users click a button, link, or image. 
In our prompt, the content of \shpd{the} ``onclick'' property refers to the functionality of the \textit{UI states} that will be accessed after clicking this element, which \shpd{is} most relevant to completing the $S_i$. Algorithm \ref{algo:prompt-augmenting} shows how to augment prompt with $\{S_1, S_2,...,  S_k\}$ and app memory $M$.


\begin{algorithm}
\footnotesize
	\caption{Prompt Augmentation}
	\label{algo:prompt-augmenting}
	\begin{algorithmic}[1]
		\Require Current user task $T$, k most similar simulated tasks in the app memory $\mathcal{S}=\{S_1, S_2... S_k\}$, App Memory $M$ 
		\Ensure Final GUI state after completing the task $T$
		
		\Function{Online-Main}{}:
			\State $Guide \gets GenerateGuide(M, \mathcal{S})$
			
			\While{$T$ not completed}
				\State $UI \gets$ Current GUI of SmartPhone
				\If{$UI \in Guide.UIs$}
					\State $UI.elements.hint \gets Guide.UI.elements.Function$ 
				\EndIf
				\State $Prompt \gets PromptGenerator(T, UI, History)$
				\State $Action \gets LLM(Prompt)$
                    \State $TaskExecutor.execute(Action)$
				\State $History.insert(Action)$
			\EndWhile \\
		\Return $Current \ GUI$
		\EndFunction\\
		
		\Function{GenerateGuide}{$M, \mathcal{S}$}: 
			\For{each $S_i$ of $\mathcal{S}$}
				\For{each $UI_i^{j}$ of $M.Simulated\_Task\_Table\{S_i\}$}
                        \State $Guide.UI_i^{j}.Element_i^{j}.hint \gets M.Function\{UI_{i}^{j+1,j+2...}\}$
				\EndFor
			\EndFor \\
		\Return $Guide$ 
		\EndFunction
	\end{algorithmic}
\end{algorithm}

Take the task shown in Figure \ref{fig:motivation} as an example. Given the task \textit{``Remove all the events in the calendar''}, \name can retrieve in the app memory and find the simulated task of the \textit{``delete all events''} button in GUI 3 to be a relevant task. Additionally, \name can find that clicking \textit{``more options''} and \textit{``settings''} in GUI 1 and GUI 2 can lead to the target button. 
Therefore, if the current UI screen is GUI 1, 
the HTML description of \textit{``more options''} in GUI 1 will change from
``<button label=`More options'></button>'', to
``<button label=`More options' onclick=\textit{`navigate to GUIs that can: 1.add contact holidays and anniversaries, import and export events, manage settings, 2.Delete all events in the app, manage event reminders, etc.}'></button>''.


\subsubsection{Tuning Local LLM with App-specific Data}
\label{sec:finetuning}
\name can also utilize smaller local LLMs (\eg Vicuna-7B \cite{vicuna2023}) to make decisions, as a cost-effective alternative to larger on-cloud LLMs (\textit{e.g.} GPT-3.5 \cite{chatgpt}).
However, the reasoning ability of these smaller LLMs is weaker than on-cloud LLMs, leading to a noticeable decrease in accuracy. It is observed that local LLMs still exhibit suboptimal performance even with the prompt augmentation methods introduced in \S \ref{sec:prompt_augment}. Researchers have found that fine-tuning using domain-specific data is an effective way to improve small LLM's abilities \cite{vicuna2023, alpaca}. Therefore, we can augment smaller LLMs by fine-tuning using app-specific data.

A key challenge in our scenario is how to generate high-quality (question, answer) pairs to fine-tune the LLM.
A naive way is to directly synthesize these data pairs from the simulated task table of the app memory. For a simulated task $S$, the memory generator records a sequence of UI states $\{U_1, U_2, ..., U_k\}$ and the UI elements $\{e_1, e_2, ..., e_k\}$ to complete it. 
We can directly generate k data pairs $(q_i, a_i)_{i=1}^k$ based on this record.
Specifically, $q_i$ is a prompt  generated based on the task $S$, previous UI actions $\{A_1, A_2, ..., A_k\}$ (where the target elements are $\{e_1, e_2, ..., e_k\}$ and the action type is \textit{click}), and the current UI state $U_i$. Then, the description of the action $A_i$ can be the answer $a_i$.
The rationale behind this approach is that: Based on the generation process of the app memory, we already know that when transitioning from interface $U_i$ to complete task $S$, action $A_i$ needs to be performed. Therefore, the correct answer of which action to choose given the state $U_i$ should be $A_i$.

However, the answers generated in this way only include \textit{<target element, action type, value>}, lacking detailed information or context. Thus, it is difficult for the local LLM to learn how to choose the correct action based on the prompt. 
If we include the reasons for choosing the target action in the answers, it will enhance the local LLM's understanding and enable it to learn how to reason based on the current task and UI \cite{LLMdistill}. 
Thus, we can ask larger LLMs (such as GPT-4 \cite{openai2023gpt4}) to answer the reason why $A_i$ is chosen to complete task $S$, and prompt it to reason in a step-by-step manner like a Zero-shot Chain-of-Thought (0-shot CoT) \cite{zero-shot-cot}. The prompt sent to the larger LLM is mainly the same as Figure \ref{fig:prompt}. Additionally, we provide the correct action to choose $A_i$, and prompt the LLM to reason about the correct action by changing the ``output requirements'' part to the following format:

\textit{Your answer should always use the following format:
1. Completing this task on a smartphone usually involves these steps: <?>.
2. Analyses of the relations between the task and the previous UI actions and current UI state: 
3. Based on the previous actions, is the task already finished? <Y/N>. The next step should be <?/None>.
4. Can the task be proceeded with the current UI state? <Y/N>. Fill in the blanks about the next one interaction: - id=<id number> - action=<tap/input> - input text=<text or N/A>.
}

The answer to the above questions can be used as the answer in the (question, answer) pair for fine-tuning the local LLM.
The thinking and reasoning data generated by these larger LLMs contains rich information and knowledge. Using it as answers to fine-tune smaller LLMs can enable it to mimic the emergent reasoning abilities of the large model.
Besides leveraging the knowledge from larger LLMs, fine-tuning LLMs with app-specific data also has the bellow two advantages: 
(i) Learning from the UTG and incorporating the insights gained from it. 
(ii) Let smaller LLMs generate answers that adhere to the desired format instead of unrestricted formatting in the answers.

\subsection{Multi-granularity Query Optimization}
\label{section: query_optimization}
We observe that the primary source of overhead in \name arises from querying LLMs. 
Consequently, reducing the frequency of LM queries for each task will result in a reduction of \name's overhead.
Additionally, as a more granular approach, pruning unnecessary tokens in the prompt, we can effectively decrease the computational cost of LLM.

\subsubsection{Pruning Tokens by Merging Functionally Equivalent Elements.}
The HTML statement of UI described in \S \ref{section: prompting} contains a lot of redundant information, which will increase the number of tokens and cause the LLM to overlook the most useful information. Therefore, We adopt two techniques to reduce the length of the text:
First, we prune the elements without any visual or textual information (such as background or container items). 
Second, we merge functionally equivalent UI elements into one element and separate the originally different elements with a ``<br>'' delimiter, which means a line-break-like spacing in HTML. 
We merge UI elements based on two rules: (i) Based on UTG: If operating on these two UI elements leads to the same interface, we combine them into a single component. Specifically, if the starting and ending points of two edges representing actions in the UTG are the same, we merge the components they operate on. 
(ii) Based on UI tree analysis: We merge the non-interactive (plain text or image) UI leaf nodes sharing the same interactive ancestor (button, checkbox, text field, etc.) in the UI tree. 
For example, in the GUI screenshot shown in Figure \ref{fig:prompt}, ``Alarms'' and ``0 items'' are two single plain-text nodes in the GUI tree that have a common clickable ancestor. 
Thus, we can merge them into an HTML statement: ``<button id=5>Alarms<br>0 items</button>'' instead of two single statements ``<button id=5>Alarms</button>'' and ``<button id=6>0 items</button>''.


\subsubsection{Reducing Query Times by Shortcuts and GUI Merging}

GUI merging is to include several GUI states into one prompt if LLMs need them all to make decisions. 
The automatic scrolling introduced in \S \ref{sec:html_desc} can accomplish this by skipping the intermediate steps like ``scroll down''. 
Without automatic scrolling, \name has to query LLMs at least twice to touch an element within the GUI after swiping, involving both scrolling and clicking. After merging the scrolled UIs into one prompt, we only need to call LLMs once and get the action \textit{``Scroll down to Button A and touch it''}. 

The shortcut is to execute simple actions directly with the help of the app memory. Although some steps are crucial and require a large model to make decisions, others are straightforward and do not require it. So if we can identify steps that are simple enough so that a local embedding model \cite{instructor-xl} can make decisions, we can reduce the number of queries. Specifically, let $T$, $E$, and $\{S_1, S_2, ...\}$ denote the user task, the embedding model, and the simulated tasks respectively. If we find $sim({E}(S_k), {E}(T))>\gamma$ where 
$S_k=\operatorname*{arg\, max}_{S_i \in \mathcal{S}} sim({E}(S_i), {E}(T))$, 
then $S_k$ is very similar to $T$, and accomplishing $S_k$ is straightforward because we have a series of actions $\{A_k^1, A_k^2, ...\}$ in the app memory that navigate from the initial UI state to $S_k$. Thus we can perform $S_k$ by the task executor without calling LLM. 
$\gamma$ is a hyper-parameter, the larger the value of $\gamma$, the stricter our criteria for selecting similar simulated tasks become.
We observe that even if the shortcut navigates to UI states unrelated to the task, LLM is still able to identify issues and quickly navigate to the correct UI states.

%% file: tex/implementation.tex
\section{Implementation}
\label{sec:implementation}


We implement \name using Python and Java.  
The local LLM Vicuna \cite{vicuna2023} is fine-tuned using PyTorch.

\textbf{Identifying Risky Actions.}
Some actions may potentially alter local or server data, or cannot be undone once performed. These actions are considered risky and require user confirmation before being executed by the agent.
For example, before calling a contact, \name needs to first prompt the user to verify the correctness of the action. If the user notices any errors in the number about to be dialed, they can manually make the necessary modifications.
\name accomplishes this by prompting the LLM to identify risky actions, \textit{i.e.} appending the sentence \textit{``If this action potentially leads to a change of user data or server state that requires user confirmation, please answer requires\_confirmation=Yes)''} to the prompt.
In addition, \name also utilizes key phrases on the UI, such as ``warning'', to further identify potentially risky actions.

\textbf{Eliding Private Information.}
We add a privacy filter that can mask the private information in the query. During online processing, it runs a Personal Identifiable Information (PII) scanner \cite{pii_detector} that can detect sensitive information in the prompt, including name, phone number, email address, etc. This personal information is replaced with non-private words (\eg ``<name>''$\rightarrow$``Alice'') before sending the prompt to the cloud. After receiving the response from LLMs, \name maps the special words back to the original ones before parsing actions.

%% file: tex/experiment.tex
\section{Benchmark}
\label{sec:benchmark}



We introduce \datasetname, an Android Task Automation benchmark suite designed to evaluate the performance of end-to-end mobile task automation systems.
\datasetname consists of 158 high-level tasks extracted from 13 popular apps. What sets our benchmark apart is that it not only provides tasks and corresponding GUI action traces but also offers the exploration memory and environment for the underlying apps. 
Agents can actively interact with the environment during the offline stage, gathering information about the apps and recording UTGs.
All 13 apps used to collect the tasks are installed, granted necessary permissions, and can reproduce the GUI action traces in our environment.
We will release the environment in the form of an Android Virtual Machine Snapshot, allowing researchers to restore the exact environment in which we collected our data.
While previous benchmarks \cite{seq2act, motif, metagui} also provide tasks and corresponding actions, they lack a reproducible environment. However, with the emergence of LLM-powered task automation methods \cite{gorilla, react}, which often require dynamic information about the environment for decision-making, our benchmark offers greater convenience for evaluating the performance of autonomous agents on mobile phones.

\begin{figure}
    \centering
    \includegraphics[width=0.48\textwidth]{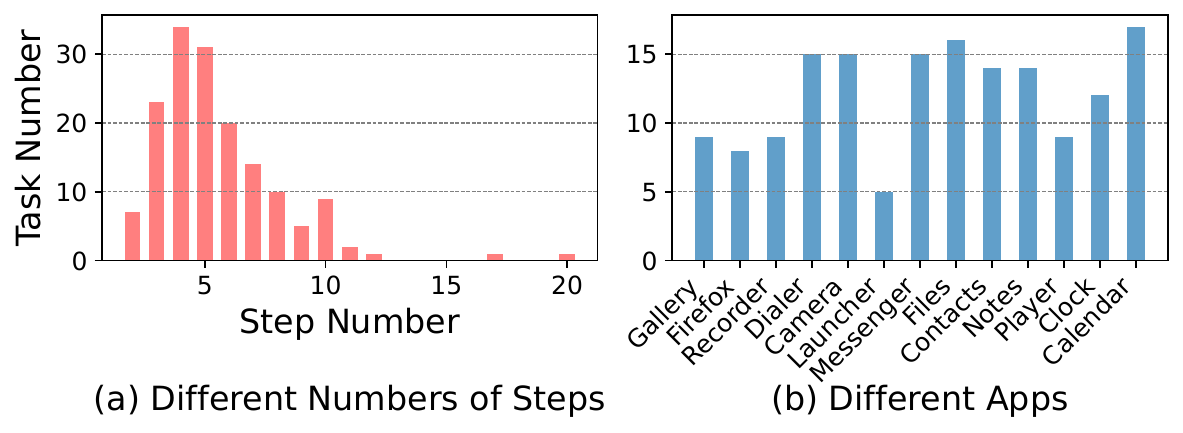}
    \vspace{-0.8cm}
    \caption{The distribution of tasks in \datasetname across different numbers of steps (a) and different apps (b).}
    \label{fig:task_destribution}
    \vspace{-0.4cm}
\end{figure}

We develop a system for collecting datasets that can interact with Android smartphones. The selected apps primarily consist of common mobile tools (such as contacts, dialer, camera, calendar, etc.) from F-Droid, a free and open-source app platform.
For each app, we ask annotators to provide a list of 5-15 tasks described in natural language. To complete each task, annotators interact with the smartphone through a desktop computer in an iterative manner.
During each iteration, the system displays the smartphone's user interface (UI) in its current state to the annotator, along with a list of available actions. Annotators can also directly observe the actual state of the smartphone. They can choose an action from the following options: 1. Touch <Button ID>, 2. Input <input text> to <EditBox ID>, 3. Swipe <Scroller ID> <direction> in the terminal.
The distribution of tasks is shown in Figure \ref{fig:task_destribution}.

\section{Evaluation}
\label{sec:experiment}


We conduct experiments to examine the accuracy and cost of \name in mobile task automation.

\subsection{Experimental Setup}
\label{eval:setup}
\textbf{Dataset.}
We mainly evaluate \name on \datasetname (mentioned in \S\ref{sec:benchmark}). We also utilize MoTiF \cite{motif} dataset to train the baseline methods and fine-tune the LLMs. MoTiF \cite{motif} is a large-scale mobile app task dataset with more than 4.7k tasks (excluding tasks without valid demonstrations). It also provides the screenshot and the tree-based representation of the GUI screens that annotators interacted with when completing these tasks, but lacks the exploration environment of the apps.

\textbf{Hardware.}
We evaluate the end-to-end performance of \name on a OnePlus ACE 2 Pro with 8 3.2 GHz ARM-based cores (Snapdragon 8 Gen2 CPU) and Adreno™ 740 GPU. \wh{The local LLM Vicuna-7B \cite{vicuna2023} is deployed on the smartphone based on Machine Learning Compilation for LLM (MLC LLM) \cite{mlc-llm}. 
Additionally, it is deployed on an edge server equipped with 1 NVIDIA 3090 24G GPU to assess inference latency in an edge computing context. }The Vicuna-7B model is fine-tuned on an 8$\times$ A100 80GB server for about 4 GPU hours.

\textbf{Baselines.}
We choose META-GUI \cite{metagui} and an existing LLM-based design for UI task automation \cite{talking_with_ui} (referred to as LLM-framework) as our main baselines.
META-GUI \cite{metagui} is a training-based conversational agent on mobile GUI that can accomplish various tasks. We train it on the MoTiF \cite{motif} dataset.
LLM-framework \cite{talking_with_ui} is an LLM-based framework that enables diverse language-based interactions with mobile UIs.
We also implement two relatively simple baselines, random performer (randomly selecting one UI element within each UI screen) and similarity-based (selecting the UI element that is semantically closest to the task using a SOTA embedding model \cite{instructor-xl}) performer. 

\textbf{Metrics.}
Given a sequence of UIs $\{U_1, U_2, ..., U_n\}$ in which human annotators performed actions $\mathcal{A}=\{A_1, A_2, ..., A_n\}$ to complete a task $T$, if one agent can make a sequence of decisions $\hat{\mathcal{A}}=\{\hat{A}_1, \hat{A}_2, ..., \hat{A}_n\}$ on $\{U_1, U_2, ..., U_n\}$, we use below two metrics to measure its performance:

(i) \textbf{Action Accuracy}: The ratio of the action $\hat{A}_i$ matching the ground-truth actions $A_i$, namely $P(\hat{A}_i=A_i)$. One action is right only if the target UI element and input text (``null'' if there is no need to input) are both right. This metric reflects the ability of the agent to make correct decisions based on the available information.

(ii) \textbf{Completion Rate}: The probability of completing all the actions in one sequence correctly, namely $P(\hat{\mathcal{A}}=\mathcal{A})$. This metric reflects the probability of the agent being able to consistently and successfully complete a task. 

\subsection{Action Accuracy}
\label{eval:action_selection}
\begin{table*}[htbp]
    \centering
    \caption{Action accuracy of \name and baselines on \datasetname. Rand: Randomly selecting actions, Sim: Similarity-based action prediction, LLM-F: LLM-framework \cite{talking_with_ui}, Complete: Determining completion.}
    
    \vspace{-0.3cm}
    \resizebox{.8\textwidth}{!}
    {
        \begin{tabular}{cccccccccc}
        \toprule
        & & & & \multicolumn{2}{c}{\textbf{Vicuna-7B}} & \multicolumn{2}{c}{\textbf{GPT-3.5}} & \multicolumn{2}{c}{\textbf{GPT-4}}\\
        \cmidrule(r){5-6} \cmidrule{7-8} \cmidrule(l){9-10} 
        \textbf{Action} & \textbf{Rand} & \textbf{Sim} & \textbf{MG}   & LLM-F & \name & LLM-F & \name & LLM-F & \name \\
        \midrule 

          Click          & 2.3\% & 35.1\% & 25.3\%  & 15.2\%      & 74.5\%   & 58.1\%       & 72.1\%   & 65.4\%    & 91.2\% \\
          Input          & 0     & 0      & 0       & 0           & 40.0\%   & 5.0\%        & 62.5\%   & 27.5\%     & 82.5\% \\
          Scroll         & 2.5\% & 0      & 0       & 8.2\%       & N/A      & 0            & N/A      & 0.6\%    & N/A \\
          Complete       & 2.5\% & 0      & N/A     & 4.4\%       & 5.7\%    & 0            & 41.8\%   & 0        & 93.7\% \\
        \midrule
         Overall         & 2.3\% & 20.8\%  &22.4\%  & 11.3\%     & \textbf{57.7\%}  & 34.7\%       & \textbf{65.1\%}   & 54.5\%    & \textbf{90.9\%} \\
        \bottomrule
        \end{tabular}
    }
    \label{tab:app_action_acc}
\end{table*}

We first evaluate the action accuracy of \name. The open-sourced LLM Vicuna-7B \cite{vicuna2023} \shpd{is fine-tuned} using the generated app-specific data, as mentioned in \S \ref{sec:finetuning}. For the \shpd{closed-source} LLM such as GPT-3.5 \cite{chatgpt} and GPT-4 \cite{openai2023gpt4}, which can not be fine-tuned directly, we augment them with automatically generated app memory, as mentioned in \S \ref{sec:prompt_augment}. 
\wh{The temperature of the LLMs is set to a lower value of 0.25 to encourage creativity while preventing it from being overly random. }
The action accuracy of \name and baselines is listed in Table \ref{tab:app_action_acc}. 
\name outperforms baselines on every action type, resulting in an overall accuracy improvement of 37.6\%. 
Among all the actions, clicking is the simplest, only requiring the decision of the element ID. On the other hand, scrolling and inputting necessitate specifying the direction or value of the UI element, and determining completion entails considering all previous actions.
It is also observed that with the LLM going larger, LLM-based methods outperform the model trained from scratch \cite{metagui}. This is because the model has only been exposed to apps and tasks from specific datasets \cite{motif}. 
Thus, it will not perform well on new apps and tasks in the \datasetname. However, by accumulating sufficient prior knowledge and incorporating our memory integration, LLMs can engage in rational reasoning on how to solve problems on new apps.
For scrollable UIs, \name will first browse and traverse all the components on the screen, eliminating the need for the ``scroll'' action. From the scroll accuracy of the baseline, it is observed that the probability of the agent actively selecting this action is very low. Thus, browsing and traversing first can improve the overall accuracy of the agent. 

The reason \name outperforms baselines is: (i) \name prunes and merges UI elements, which reduces the action space (from 36.4 to 13.2 choices per GUI state on average). (ii) The exploration-based memory can enhance the LLM with domain-specific knowledge about the apps, which will be detailed in \S \ref{eval:ablation}. (iii) The output format of the fine-tuned model is aligned more closely with the format requirements specified in the output requirements. If the output is not standardized, the task executor would be unable to extract or recognize the element ID and action.



We further analyze why and how \name fails on some steps. We randomly sample 20 failure cases by \name (\shpd{using} GPT-4 \cite{openai2023gpt4} as the LLM), and categorize 3 typical failure causes, as explained below:
1. Multiple Correct Choices. In certain cases, there can be multiple valid ways to complete a task.
Annotators may not be able to exhaustively list all the possible ways to complete a task, and if the agent attempts a different approach than what the annotators specified, it may be deemed incorrect.
2. Unable to accurately determine if the task has been completed. Sometimes \name mistakenly considers a task completed when it detects the presence of a specific UI element. 
3. \shpd{Lack of} understanding of the GUI. \name occasionally overlooks important information or hints in the UI and makes decisions based on its prior experience. For example, in the task \textit{``open the camera and record a short video, name it `test.mp3' ''}, the agent only needs to input `test' into the ``name'' box. This is because the GUI indicates that the file extension `.mp3' is already displayed in the ``file type'' box. However, \name still selects `test.mp3' as the input to the ``name'' box.

\subsection{Task Completion Rate}
\label{eval:task_completion}

\begin{figure}
    \centering
    \includegraphics[width=0.48\textwidth]{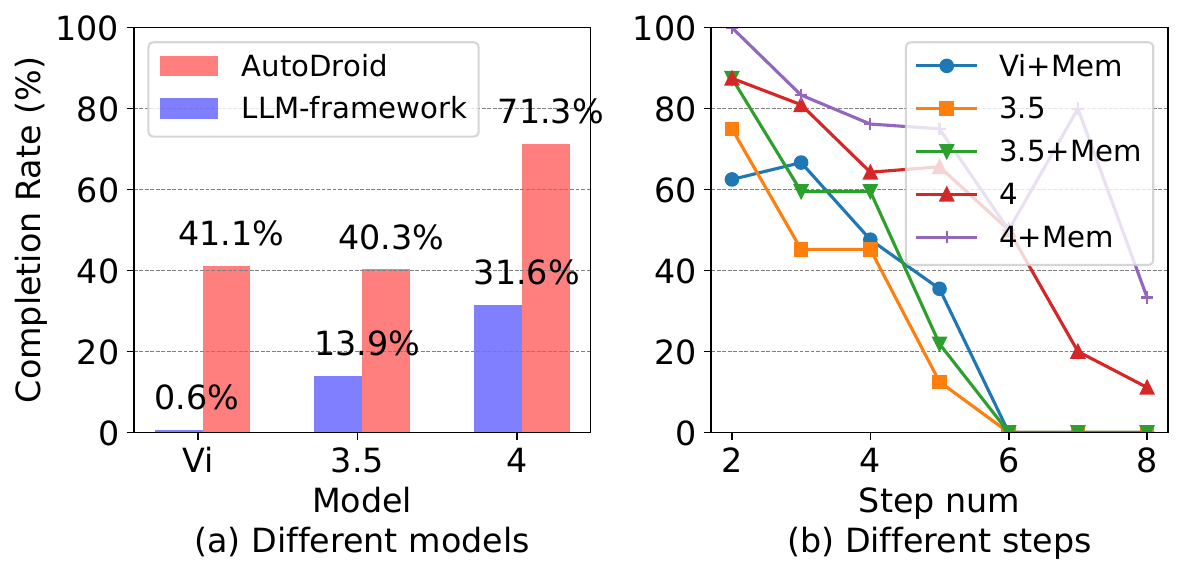}
    \vspace{-0.8cm}
    \caption{Task completion rate of \name with different LLMs and with \shpd{varying numbers} of steps. Vi: Vicuna-7B, 3.5: GPT-3.5, 4: GPT-4.}
    \label{fig:completion_rate}
    \vspace{-0.2cm}
\end{figure}

The Task Completion Rate of \name and LLM-framework \cite{talking_with_ui} is shown in Figure \ref{fig:completion_rate} (a). Note that we do not include the completion determination step for clear comparison. \name outperforms baseline by 40.5\%, 26.4\%, and 39.7\% for Vicuna-7B, GPT-3.5, and GPT-4 respectively. We also show the completion rate of \name with and without memory augmentation in Figure \ref{fig:completion_rate} (b). As the number of steps increases, the overall completion rate decreases. This is because (i) the probability of each step being executed correctly decreases. (ii) Tasks that involve multiple steps often have multiple approaches to completion (e.g., creating a new contact by entering either the name or the phone number first). However, human annotators typically only annotate one approach, which can lead to the model's solution being mistakenly judged as incorrect. The actual completion rate in the real-system can be higher than the reported results, but we do not include real-system results since determining task completion can be ambiguous.

\subsection{Ablation Study}
\label{eval:ablation}
\subsubsection{Memory Injection}
\begin{figure}
    \centering
    \includegraphics[width=0.45\textwidth]{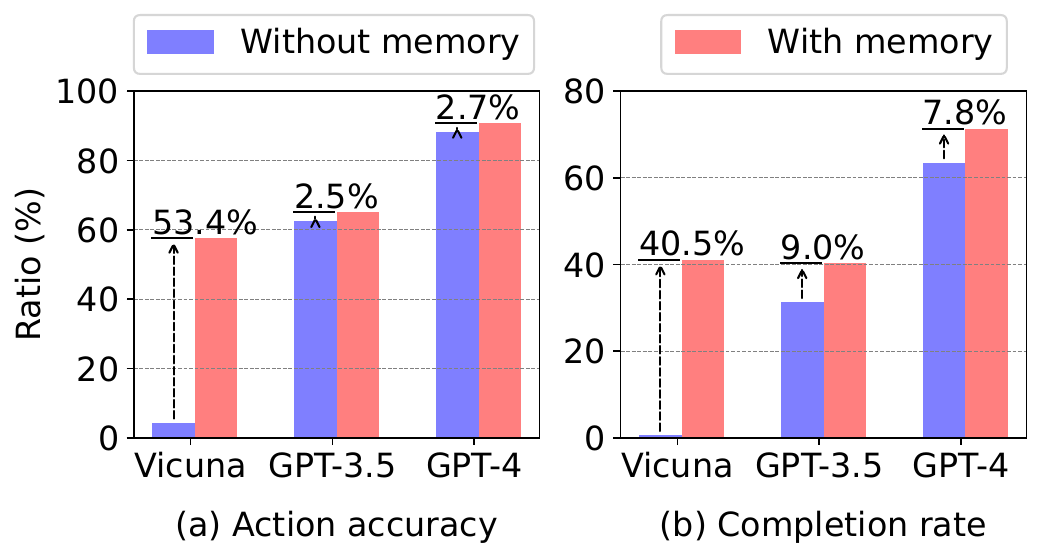}
    \vspace{-0.4cm}
    \caption{Action accuracy and task completion rate of \name, with and without memory augmentation.}
    \label{fig:ablation_memory}
    \vspace{-0.2cm}
\end{figure}
The action accuracy and task completion Rate of \name with and without memory is shown in Figure \ref{fig:ablation_memory}. We can observe that the improvement in overall completion rate is much higher than the improvement in single-step accuracy. This is because the introduction of memory allows LLMs to make crucial single-step decisions (such as the example in Figure \ref{fig:motivation}). Although these critical steps account for a small proportion of all the action steps, they are essential for successfully completing certain tasks.
Moreover, it can be observed that smaller models benefit more from the inclusion of memory in terms of task completion rate. This is because smaller models possess less prior knowledge, thus requiring more guidance from application-specific knowledge. However, even for larger models, the incorporation of memory remains meaningful. Limited \shpd{model capacity} cannot store the vast and ever-growing knowledge present in the world, making it difficult to stay updated on the evolving usage patterns of new applications. Therefore, automatic exploration and recording of their usage patterns play a crucial role in enabling LLMs to effectively use applications.

\subsubsection{Zero-shot Chain of Thought Fine-tuning}

\begin{table}
 \caption{Action accuracy and completion rate of \name based on Vicuna-7B with different fine-tuning techniques. Original: Vicuna-7B without fine-tuning. CoT: Fine-tuning with zero-shot chain-of-thought. Mo: Incorporating a small portion of MoTiF \cite{motif} dataset for fine-tuning.} 
    \vspace{-0.3cm}
    \centering
    \resizebox{.48\textwidth}{!}
    {
        \begin{tabular}{cccccc}
		\toprule
            Metric      &  Original     &  \name      & No Mo  &  No CoT   & No Mo\&CoT \\
		\hline
            Action        &   11.3\%  &   57.7\%   &   51.9\%    &  20.6\%  &  51.9\%   \\
		  Completion    &   0.6\%   &   41.1\%   &   29.8\%    &   0.6\%  &  31.6\%   \\
            
		\bottomrule
	\end{tabular}
    }
 \label{tab:cot}
\end{table}

The action accuracy and task completion rate of \name based on Vicuna-7B \cite{vicuna2023} fine-tuned with and without Zero-shot Chain-of-Thought (0-shot CoT) \cite{zero-shot-cot} is shown in Table \ref{tab:cot}. Since the app memory automatically generated by \name only contains clicking and checking actions, the LLM fine-tuned merely on the app memory is poor on inputting and adjusting whether the task has been completed.
Hence, we incorporate a small portion of manually annotated data for fine-tuning. Specifically, we add only the input and completion judgment data from MoTiF dataset \cite{motif} into the app memory dataset. Note that the app and task in the MoTiF dataset \cite{motif} are unrelated to our dataset. Thus, adding this portion of data will not result in any test data leakage. It simply enables the model to learn to input and to determine the task's completion.

Vicuna-7B \cite{vicuna2023} fine-tuned with Zero-shot Chain-of-Thought data generated by app memory mixed with a small portion of MoTiF \cite{motif} (\textit{\name}) can achieve 57.7\% action accuracy and 41.1\% completion rate on \datasetname, with an input accuracy of 40.0\%.
Without MoTiF \cite{motif} data (\textit{``No Mo''}), the fine-tuned model can achieve 51.9\% action accuracy, and the inputting accuracy is 0\%. 
We observe that when there are no CoT and no MoTiF data (\textit{``No Mo\&CoT''}), the fine-tuned LLM can achieve a high accuracy rate with simple click actions, and it can generally handle tasks that involve only clicking. However, once the MoTiF dataset is introduced (\textit{``No CoT''}) to teach the LLM additional types of actions (such as input and task completion judgments), the LLM is heavily misled by the completion of judgment tasks. As a result, it outputs a significant number of ``task completed'' instead of selecting actions correctly. Consequently, the action accuracy drops from 51.9\% to 20.6\%.
\vspace{-0.3cm}
\subsection{Cost Analysis}
\label{eval:overhead}

\begin{figure}
    \centering
    \includegraphics[width=0.48\textwidth]{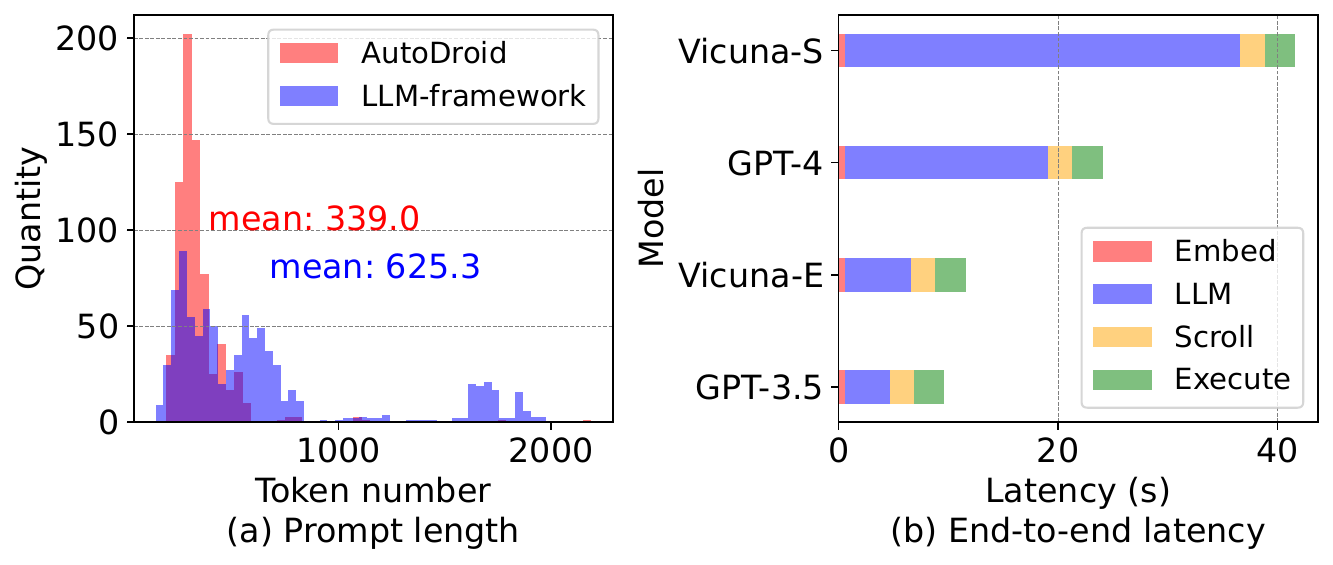}
    \vspace{-0.7cm}
    \caption{Overhead of \name and LLM-framework \cite{talking_with_ui}. Left: The number of prompts with different token counts. Right: The component of per-step latency of \name based on 3 LLMs respectively. \wh{Vicuna-E: Vicuna-7B deployed on the edge server. Vicuna-S: Vicuna-7B deployed on the smartphone.} }
    \label{fig:prompt_length}
    \vspace{-0.3cm}
\end{figure}

\textbf{Runtime Cost.}
\name reduces runtime overhead by addressing two aspects: reducing the number of tokens per query and minimizing the frequency of query. In Figure \ref{fig:prompt_length} (a), we show the count for each prompt length. Our baseline \cite{talking_with_ui} \shpd{includes} only visible leaf nodes in the UI tree, and contains 625.3 tokens within each prompt on average. \name merges functionally equivalent nodes in the UI tree and further simplifies the expression of properties, reducing the token count by nearly half (339.0 on average). 
There are two main benefits: (i) Reducing token length can significantly decrease the model's inference latency. (ii) For calling on-cloud LLM API, it can reduce costs. For example, for GPT-3.5 and GPT-4, the cost can be reduced from \$0.938 and \$18.76 to \$0.509 and \$10.17 every 1000 queries respectively on average.

In Table \ref{tab:inference_lat}, \shpd{we} randomly select five baseline prompts and find the corresponding prompts optimized by \name. \wh{We measure their real latency on the Vicuna-7B \cite{vicuna2023} deployed on the smartphone as well as on the edge server. Our optimized prompt reduces inference latency by 21.3\% on average.} Note that the inference latency of LLMs on the smartphone and the edge server primarily depends on the number of output tokens. Therefore, when deploying the LLM on a mobile device, we do not require the LLM to output the Chain-of-Thought but rather output in the original manner shown in Figure \ref{fig:prompt}. In the case of \textit{$P_5$}, due to the excessive length of the baseline, it was truncated after outputting only one word, resulting in minimal inference latency.

Figure \ref{fig:prompt_length} (b) shows the component of per-step latency of \name. \wh{The Vicuna-7B model is deployed on the smartphone and on the edge server. }On-cloud GPT-3.5 and GPT-4 models are accessed by making API calls. The embedding model \cite{instructor-xl} is deployed on an edge 1080 Ti GPU with 11 GB memory.
\wh{Note that the latency in calling GPT-3.5 and GPT-4 is significantly influenced by network conditions, server load, and so on. Therefore, we \shpd{make 10 measurements to calculate the average latency}, but there still remains a considerable degree of instability. }
\wh{Calling LLMs (\textit{``LLM''}) accounts for the majority of the latency, with 42.1\%, 51.9\%, 77.6\%, and 87.1\% of the latency based on GPT-3.5, Vicuna-7B (on-server), GPT-4, and Vicuna-7B (on-device) respectively. }Therefore, reducing LLM calls can largely reduce the end-to-end overhead. Besides, Embedding the task and searching the most similar UI element  (\textit{``Embed''}) account for only 1.7\% of the overhead, and only needs to be executed once for every task. Therefore, the overheads of finding the shortcuts and memory injection are acceptable.

\begin{table}
 \caption{Per-step inference cost of \name with Vicuna-7B deployed on the OnePlus ACE 2 Pro smartphone. LLM-F: LLM-framework \cite{talking_with_ui}. $P_{1\sim5}$: Five random prompts from LLM-framework.}
    \vspace{-0.1cm}
    \centering
    \resizebox{.45\textwidth}{!}
    {
        \begin{tabular}{crrrrr}
		\toprule
		\cmidrule(r){1-2}
            Prompt length / Inference latency     & $P_1$  & $P_2$  & $P_3$  & $P_4$  & $P_5$    \\
		\midrule
            LLM-F input length (token)    &252  &401 &460  &559   &  719 \\
		  \name input length (token)    &299  &280 &233  &177   &  233 \\
            \midrule
		On device LLM-F latency (s)             &40.6  &50.2 &63.9  &64.9 &  36.0  \\
            On device \name latency (s)   &39.7  &30.8 &33.9  &39.7 &  22.7 \\
            \wh{On cloud LLM-F latency (s)}&\wh{4.4}  &\wh{5.5} &\wh{6.4}  &\wh{16.1} &  \wh{6.5} \\
            \wh{On cloud \name latency (s)} &\wh{4.2}  &\wh{8.8} &\wh{4.9}  &\wh{5.3} &  \wh{5.6} \\
            
		\bottomrule
	\end{tabular}
	}
 \label{tab:inference_lat}
 \vspace{-0.2cm}
\end{table}

We also conducted experiments on saving the number of calls based on merging GUI and shortcuts. 
On average, \name reduces LLM calls by 1.2 per task resulting in an overall decrease of 13.7\% in the total number of calls using the GUI merging technique.
Our shortcuts correctly guide LLMs in 75\% of cases. Considering only the correct shortcuts, we save 38.02\% of the number of steps, with an average savings of 1.73 steps per task.

\textbf{Offline Preparation Cost.}
For every app, it takes about 0.5-1 hour to generate the UI Transition Graph (UTG), which is then analyzed to synthesize simulated tasks based on LLMs, taking about 5-10 minutes. Finally, the simulated tasks are mapped into high-dimensional vectors by an embedding model \cite{instructor-xl} for runtime lookup, which typically takes about 10 seconds on a desktop computer. The offline preparation is a one-time process and does not need to be performed again at runtime.

\subsection{Influence of Security/Privacy Filtering}
\label{eval:security_privacy}

We ask the annotators of \datasetname to also determine whether each action could potentially change the state of the user or the app. If so, we consider the action to be risky and prompt the user to confirm whether to proceed with \shpd{the action}. We evaluate \name's accuracy in detecting risky actions in five apps that may contain risky actions (contacts, dialer, SMS messenger, clock, and calendar). We consider risky actions as positive examples and \name achieved a precision of 75.0\% and a recall of 80.5\%. We further show the influence of adding privacy information replacing and security confirmation into the prompt in Table \ref{tab:dangerous_action}. When privacy replacement and security confirmation are added, a decrease in accuracy and completion rate can be observed, which is acceptable.

    



\begin{table}
 \caption{Action accuracy and completion rate of \name based on GPT-4 with privacy information \shpd{replacement} and security confirmation on 5 apps in \datasetname. Priv: Privacy information \shpd{replacement}, Sec: Security confirmation.} 
    \vspace{-0.1cm}
    \centering
    \resizebox{.35\textwidth}{!}
    {
        \begin{tabular}{ccccc}
		\toprule
		\cmidrule(r){1-2}
  
            Metric            & Original  & +Priv  & +Sec  & +Priv\&Sec    \\
		\midrule
		  Acc               &  92.9\%   &  89.9\%  &  89.9\%   &  89.9\%  \\
		Completion        &  75.4\%   &  69.9\%  &  68.5\%   &  69.9\%  \\
		\bottomrule
	\end{tabular}
	}
 \vspace{-0.4cm} 

 \label{tab:dangerous_action}
\end{table}





%% file: tex/related_work.tex
\section{Related Work}\label{sec:related_work}

\wh{\textbf{UI Understanding and Automation. }\shpd{There has been growing interest in using machine learning techniques to comprehend and summarize user interfaces, enabling use cases such as accessibility and task-oriented bots.} Key areas of research include: 1) Semantic analysis of GUIs to summarize functions \cite{vut, spotlight}, interpret UI elements' purposes \cite{li-etal-2020-widget, screen_recognition}, and address user questions related to the GUI \cite{screen2words, talking_with_ui}. It is crucial for various interaction tasks such as UI automation and accessibility.
2) Mapping user instructions to UI elements \cite{seq2act, metagui, kite}. These methods aim to to select the most relevant GUI elements for given tasks. 
3) Mobile UI task automation \cite{responsible_task_automation,wen2023droidbot-gpt}. These methods build agents to complete tasks for users by performing actions on the GUI. 
\name, on the other hand, leverages the UI transition memory to complete complex, multi-step tasks on smartphones. The memory can help agents to understand richly informative UIs and the usage of apps, and augment the LLMs in reasoning and planning.
After the first release of \name, there were various LLM-based UI agents proposed, which had been comprehensively summarized in a recent survey \cite{li2024personal}.}

\textbf{Augmented LLM.}
Although LLMs excel in tasks like question answering and text generation, they are still constrained by the information they can store in their fixed set of weights and context length. Therefore, researchers are augmenting LLMs with different tools, such as web browser \cite{webgpt, mind2web}, APIs \cite{chamelon, gorilla}, and other DNN models \cite{hugginggpt}. 
Unlike existing approaches that often depend on public APIs, our method does not require custom APIs, which are uncommon in mobile applications.


%% file: tex/discussion.tex
\section{Discussion}\label{sec:discussion}
\textbf{Randomness of LLMs.} We can set the `temperature' hyperparameter to 0 for consistent responses. But setting temperature too small will \shpd{inhibit} innovative answers,  \shpd{thereby potentially reducing} the performance of our system. In our experiments, we set the temperature to 0.25. And we observe a 2.1\% accuracy \shpd{reduction} when we set the `temperature' of GPT-3.5 to 0. \shpd{Conversely}, increasing the temperature to 0.7 boosted action accuracy by 3.8\%.

\textbf{Increased latency} limits the practical use of \name. Our work could be extended by a collaborative approach between LLMs and smaller models. 
We could call LLMs only once for each task to create a guideline based on the filtered domain-specific knowledge about the app. Subsequently, smaller models could be employed to associate these guidelines with UI elements \cite{seq2act, metagui}. 
 Introducing an instruction cache could further reduce latency by storing and reusing common commands, minimizing the need for repeated LLM invocations.

%% file: tex/conclusion.tex
\section{Conclusion}
We present an LLM-powered mobile task automation system that can support arbitrary tasks without manual efforts.
Experiment results have shown that our method can achieve effective task automation, outperforming existing training-based and LLM-based baselines.
We believe that the synergy between the commonsense knowledge of LLMs and domain-specific knowledge in mobile apps can potentially bring truly intelligent and helpful personal assistants into reality.
\section*{Acknowledgement}
\label{sec:acknowledgement}
This work is supported by the National Key R\&D Program of China (No.2022YFF0604501), NSFC (No.62272261), and Tsinghua University (AIR)--AsiaInfo Technologies (China) Inc. Joint Research Center.